\documentclass[letterpaper, 10 pt, conference]{ieeeconf}  

\IEEEoverridecommandlockouts                              

\usepackage{graphics} 
\usepackage{epsfig} 
\usepackage{amsmath} 
\usepackage{amssymb}  

\title{\LARGE \bf
A hierarchical framework for collision avoidance in robot-assisted minimally invasive surgery}

\author{Jacinto Colan$^{1}$, Ana Davila$^{2}$ and Khusniddin Fozilov$^{1}$ and Yasuhisa Hasegawa$^{2}$
\thanks{$^{1}$\quad
Department of Micro-Nano Mechanical Science and Engineering, Nagoya University, Furo-cho, Chikusa-ku, Nagoya, Aichi 464-8603, Japan}
\thanks{$^{2}$\quad
Institutes of Innovation for Future Society, Nagoya University, Furo-cho, Chikusa-ku, Nagoya, Aichi 464-8601, Japan}
\thanks{Correspondence: {\tt\small colan@robo.mein.nagoya-u.ac.jp}}
\thanks{
This work was supported in part by the Japan Science and Technology Agency (JST) CREST under Grant JPMJCR20D5, and in part by the Japan Society for the Promotion of Science (JSPS) Grants-in-Aid for Scientific Research (KAKENHI) under Grant 22K14221.}
}

\begin{document}

\maketitle
\thispagestyle{empty}
\pagestyle{empty}

\begin{abstract}
Minimally invasive surgery (MIS) procedures benefit significantly from robotic systems due to their improved precision and dexterity. However, ensuring safety in these dynamic and cluttered environments is an ongoing challenge. This paper proposes a novel hierarchical framework for collision avoidance in MIS. This framework integrates multiple tasks, including maintaining the Remote Center of Motion (RCM) constraint, tracking desired tool poses, avoiding collisions, optimizing manipulability, and adhering to joint limits. The proposed approach utilizes Hierarchical Quadratic Programming (HQP) to seamlessly manage these constraints while enabling smooth transitions between task priorities for collision avoidance. Experimental validation through simulated scenarios demonstrates the framework's robustness and effectiveness in handling diverse scenarios involving static and dynamic obstacles, as well as inter-tool collisions.
\end{abstract}

\section{INTRODUCTION}

\begin{figure}[t]
  \centering
  \includegraphics[width=0.75\linewidth]{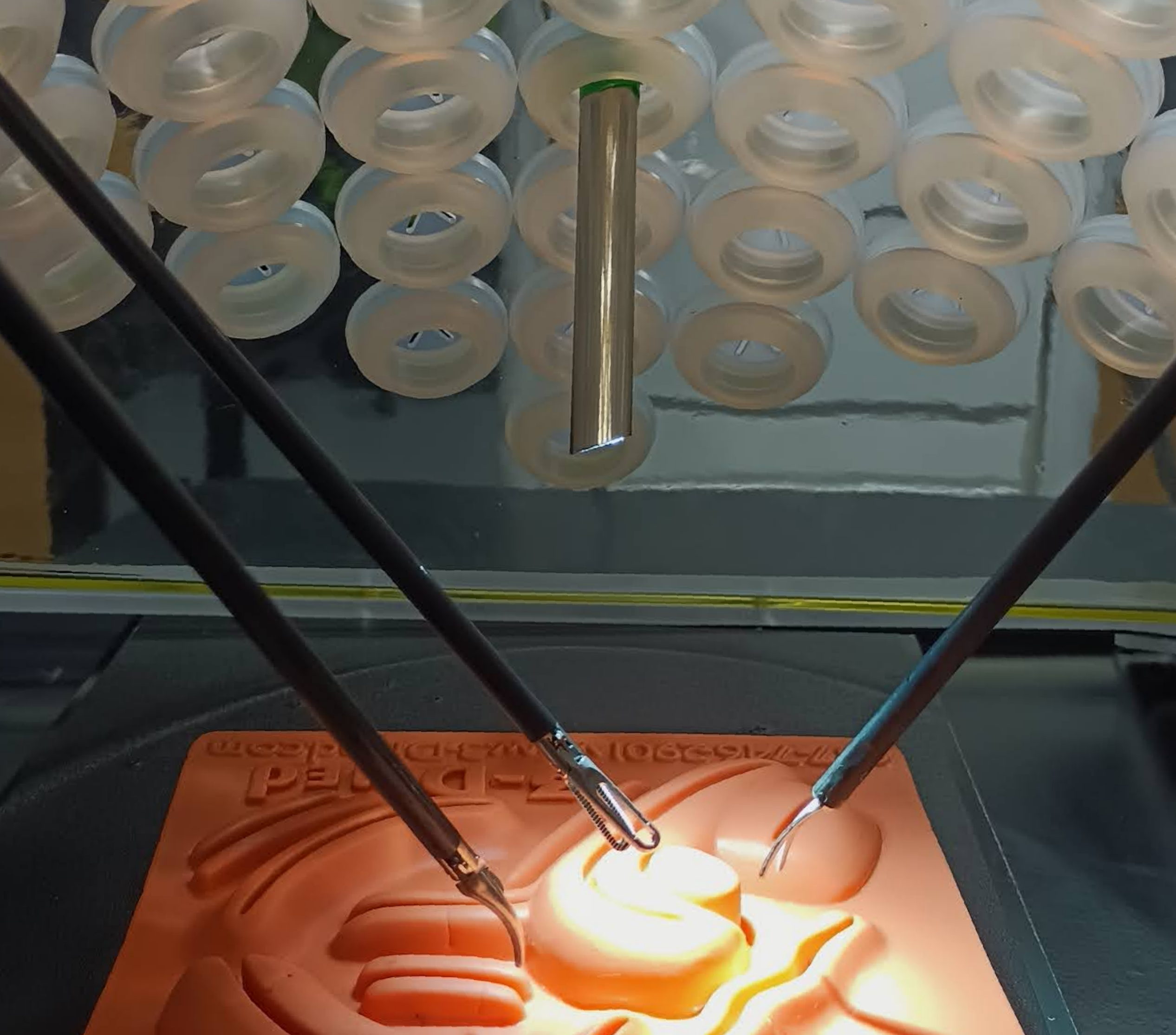}
  \caption{Multiple surgical tools are commonly used in a constrained workspace.}
  \label{fig:1}
\vspace*{-5mm}
\end{figure}

Minimally invasive surgery (MIS) offers numerous advantages over traditional open surgery, including reduced patient trauma, faster recovery times, and improved surgical results. However, it poses significant challenges in maintaining stable and safe control of teleoperated or autonomous surgical systems. These procedures typically require the concurrent operation of multiple instruments within restricted spaces, which are prone to collisions, as illustrated in Fig.~\ref{fig:1}. This is particularly evident during multi-hand surgical tasks, such as tissue resection \cite{fozilov23endoscope} or tissue triangulation \cite{liu24latent}. The presence of a remote center of motion (RCM) constraint further complicates the task, as the tool needs to pivot around the insertion point (usually the trocar). Additional constraints, such as kinematic limitations, are also important to ensure accurate tool positioning and manipulation dexterity. Therefore, robust and adaptable collision avoidance strategies are essential to handle the dynamic constraints inherent in real-time surgical environments.

Several techniques address collision avoidance in Robot-Assisted Minimally Invasive Surgery (RAMIS), including active constraints, haptic feedback, sensing-based control, and reactive motion planning. These techniques aim to prevent unintended contact between the robotic instruments and non-target tissues or between the surgical tools themselves. For instance, Marinho et al. \cite{marinho19dynamic} proposed a vector-field-inequalities method to establish dynamic active constraints between a robot and moving objects sharing the same workspace. Moccia et al. proposed a collision avoidance method based on Forbidden Region Virtual Fixtures, which render a repulsive force to the surgeon \cite{moccia20vision}. Li et al. \cite{li23dimensional} introduced a three-dimensional collision avoidance method based on three strategic vectors: a collision-with-instrument-avoidance vector, a collision-with-tissues-avoidance vector, and a constrained-control vector. Banach \cite{banach19active} developed a method to prevent surgical tool-clashing and tool-shaft collisions with delicate anatomy using elasto-plastic frictional force control. However, most of these techniques rely on simplified models or assumptions that treat the collision avoidance problem independently of other constraints found in surgical environments, such as RCM or joint limits. Beyond surgical applications, other approaches for collision avoidance have been proposed, including artificial potential fields \cite{li18algorithm}, evolutionary search \cite{mingxin10hybrid, kong20semg}, fuzzy logic \cite{zhou15descentralized}, sampling-based motion planning \cite{fozilov23toward, yu19ant}, and artificial neural networks \cite{he16adaptive}.

The challenge of handling multiple constraints with different priorities is commonly addressed using prioritized strategies. The most prevalent approach, null-space projection, is effective in managing multiple constraints with different priorities, but can encounter limitations when dealing with inequality constraints, such as joint limits, and is susceptible to local minima. In contrast, optimization-based methods can incorporate inequalities into the optimization problem and prioritize tasks by assigning weights according to their relative importance. However, they may not guarantee strict prioritization. On the other hand, Hierarchical Quadratic Programming (HQP) enables solving multiple tasks hierarchically as Quadratic Programming (QP) problems based on their priority \cite{escande14hierarchical}. HQP can efficiently handle multiple objectives and constraints while ensuring consistency.

This paper introduces a hierarchical framework for collision avoidance in surgical setups that integrates both strict and soft task priorities. It incorporates fundamental tasks and constraints commonly encountered in Minimally Invasive Surgery (MIS), including Remote Center of Motion (RCM), tracking, manipulability, and kinematic constraints.

\section{Proposed approach}

\subsection{Hierarchical framework for surgical robot motion planning}

In the context of MIS setups, several critical tasks and constraints must be addressed, including adhering to the Remote Center of Motion constraint, tracking a desired tool pose, preventing collisions with other tools or tissues, maximizing manipulability to ensure dexterity and force range, and respecting kinematic limits.

\begin{figure}[thpb]
  \centering
  \includegraphics[width=0.45\linewidth]{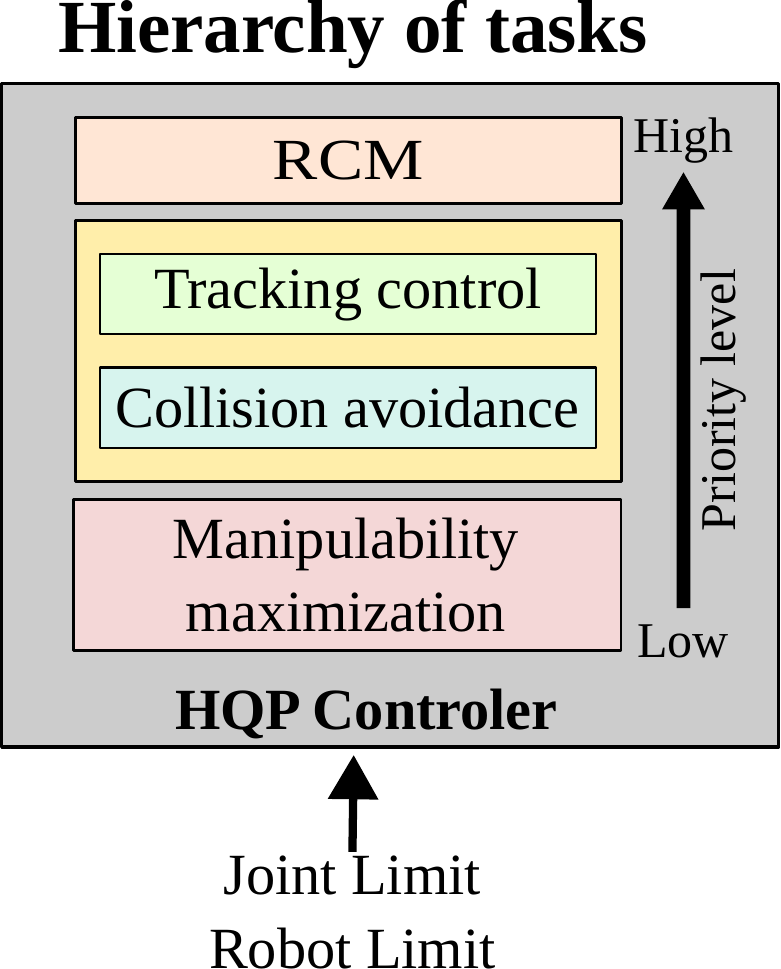}
  \caption{Hierarchy of tasks for the proposed framework}
  \label{fig:2}
  \vspace*{-3mm}
\end{figure}

To efficiently manage these tasks while considering their relative importance, a Hierarchical Quadratic Programming (HQP) based controller is proposed. This controller allows for the simultaneous handling of these constraints in real-time, with smooth transitions between hierarchies, particularly concerning collision avoidance.  The task priorities define the task hierarchy, as shown in Fig.~\ref{fig:2}, are assigned based on their significance in the robot's motion control.

\subsection{HQP controller}

To address the inverse kinematics problem (IKP) for our HQP controller, multiple tasks and constraints must be integrated and solved concurrently while respecting the hierarchy between priority levels. Additionally, tasks of equal priority can be accommodated, with the option to introduce soft restrictions by assigning weights to each task within the optimization problem. A generalized representation of the IKP problem for a priority level $p$ is as follows \cite{colan23concurrent}:

\begin{equation}
    \begin{aligned}
        \min_{\dot{q},w} \quad & \sum_{i=1}^{\eta_p} \frac{K_{t_i}}{2} || J_i \dot{q} - K_{r_i} r_i||^2_2  + \frac{K_d}{2} ||\dot{q}||_2^2 + \frac{K_w}{2} ||w||_2^2\\
        \textrm{s.t.} \quad & \qquad \qquad \qquad C \dot{q} - d_p \le w \\
    \end{aligned}
    \label{eq:3}
\end{equation}

where $J_i$ and $r_i$ represent the $i$th Jacobian matrix and the residual respectively. $K_{t_i}$ and $K_{r_i}$ are positive weights for task $i$. The slack variable $w$ is used to incorporate the inequality constraints and the squared norm of $\dot{q}$ works as a regularization term. $K_d$ and $K_w$ are positive weights for the regularization and slack terms, respectively. $C_p$ and $d_p$ are a general matrix and vector that represent the inequality constraints for the tasks with priority $p$. A Closed-loop IK scheme (CLICK) is used for regularization problems ($\dot{x} = 0$), replacing $\dot{x}$ with $K_{r_i}r_i$ to ensure robustness against singularities \cite{chiaverini97singlarity}.   

The optimization problem defined in the above equation is specified for each priority level and is solved as a Quadratic Programming (QP) problem. Hierarchy is enforced by maintaining optimality conditions between successive tasks, as proposed by Kanoun et al. \cite{kanoun11kinematic}. Higher-priority tasks are shielded from the influence of lower-priority tasks by incorporating the null-space projector operator into the optimization formulation, ensuring that lower-priority tasks do not interfere with the execution of higher-priority tasks. The optimization problem, considering these optimality conditions, is as follows:

\begin{equation}
    \begin{aligned}
        \min_{\dot{q},w} \quad & \sum_{i=1}^{\eta_p} \frac{K_{t_i}}{2} || J_i \left( N_{p-1} \dot{q} + \dot{q}_{p-1}^* \right) - K_{r_i} r_i||^2_2  + \frac{K_d}{2} ||\dot{q}||_2^2 \\
        & + \frac{K_w}{2} ||w||_2^2\\
        \textrm{s.t.} \quad & \qquad C_p \left( N_{p-1} \dot{q} + \dot{q}_{p-1}^* \right) - d_p \le w \\
        \quad & \qquad C_{p-1} \left( N_{p-1} \dot{q} + \dot{q}_{p-1}^* \right) - d_{p-1} \le w_{p-1}^* \\
        \quad & \qquad \qquad \qquad \qquad \qquad \vdots \\
        \quad & \qquad C_1 \left( N_{p-1} \dot{q} + \dot{q}_{p-1}^* \right) - d_{1} \le w_{1}^* \\
    \end{aligned}
    \label{eq:4}
\end{equation}

where $N_{p-1}$ represents the null space projector of the higher priority level, $N_0 = \mathbb{I}$ and $\dot{q}^*_0 = \mathbf{0}$. 

Each optimization problem can be solved as a QP problem, where:

\begin{equation}
\label{eq:16}
    \begin{aligned}
         x_p^* = \min_{x} \quad & \frac{1}{2} x^T Q_p x + c_p^T x \\
        \textrm{s.t.} \quad & \overline{C_p}x - \overline{d_p} \le 0
    \end{aligned}
\end{equation}

where the optimization variable is represented by $x = [\dot{q} \quad w]$, $Q=\overline{A_p}^T\overline{A_p}$ and $c=-\overline{A_{p}}^T \overline{b_p}$. The matrices $\overline{A_p}$, $\overline{C_p}$ and the vectors $\overline{b_p}$, $\overline{d_p}$ are derived from the task formulation (Eq.~\ref{eq:4}) as 

\begin{equation}
    \overline{A_p} = \begin{bmatrix} A_p N_{p-1} & 0 \\ 0 & K_w^{1/2} I \end{bmatrix}
    \quad \quad
    \overline{b_p} = \begin{bmatrix} b_p \\ 0  \end{bmatrix}
\end{equation}

\begin{equation}
    A_p = \begin{bmatrix} K_{t_1}^{1/2} A_1 \\ \vdots \\ K_{t_n}^{1/2} A_n \\ K_{d}^{1/2} I \end{bmatrix}
    \quad\ \quad 
    b_p = \begin{bmatrix} K_{t_1}^{1/2} \left( A_1 \dot{q}_{p-1}^* - b_1 \right) \\ \vdots \\  K_{t_n}^{1/2} \left( A_n \dot{q}_{p-1}^* - b_n \right) \\ \mathbf{0} \end{bmatrix}
\end{equation}

The optimal solution $\overline{x_p}^*$ is then given by

\begin{equation}
    \overline{x}_p^* = N_{p-1} x_p^* + \overline{x}_{p-1}^*
\end{equation}

where $\overline{x}_{p-1}^*$ represents the optimal solution for the optimization problem of higher priority $p-1$.

\subsection{Hierarchy of Tasks}

The hierarchy of tasks in our framework encompasses various objectives, each subject to joint and velocity constraints. These objectives include the Remote Center of Motion (RCM) constraint, tracking control, manipulability maximization, and collision avoidance. To integrate these tasks into the proposed HQP controller, we formulate them as Quadratic Programming (QP) problems.

\subsubsection{Remote center of Motion}
To ensure that the surgical tool pivots around the RCM, the optimization problem minimizes the norm of the vector $p_e$ representing the deviation from the nearest point along the tool axis ($p_{rcm} \in \mathbb{R}^3$) to the trocar point ($p_{trocar} \in \mathbb{R}^3$). The RCM task Jacobian matrix $J_{rcm(q)} \in \mathbb{R}^{1\times n}$ can be calculated as:

By differentiating the RCM deviation $||p_{rcm}||$ with respect to the manipulator joints $q \in \mathbb{R}^{n\times 1}$ , the RCM task Jacobian matrix $ J_{{rcm}_{(q)}} \in \mathbb{R}^{1\times n}$ can be calculated as

\begin{equation}
\begin{aligned}
    &J_{rcm}(q) = -\hat{p_e}^T \frac{\delta p_{rcm}}{\delta q} \\ 
\end{aligned}
\end{equation}

A formulation for computing $\frac{\delta p_{rcm}}{\delta q}$ is given in \cite{davila24realtime}. Given that the objective is to achieve no deviation from the RCM, the residual for the RCM task $r_{rcm}$ is computed as $-||p_e||$.
  
\subsubsection{Tracking control}

For the tracking task, the goal is for the end effector of the surgical tool to reach a desired pose $X_{{ee}_{des}} \in SE(3)$ from its current initial tool pose $X_{{ee}_{act}}$. The task Jacobian matrix $J_{{ee}(q)}$ is defined as $J_{ee}(q) = \frac{\delta{^B}X_{{ee}_{act}}}{\delta q}$. The tracking task's residual is computed as \cite{colan24variable}:

\begin{equation}
    r_{{ee}_{(q)}} = \log_6(X_{{ee}_{des}}X_{{ee}_{act}}^{-1})
\end{equation}
where the logarithm $\log_6 : SE(3) \rightarrow se(3)$ maps the pose from the Lie group $SE(3)$ to twists in the $se(3)$ \cite{sola18micro}.

\subsubsection{Collision avoidance}

\begin{figure}[t]
  \centering
  \includegraphics[width=0.8\linewidth]{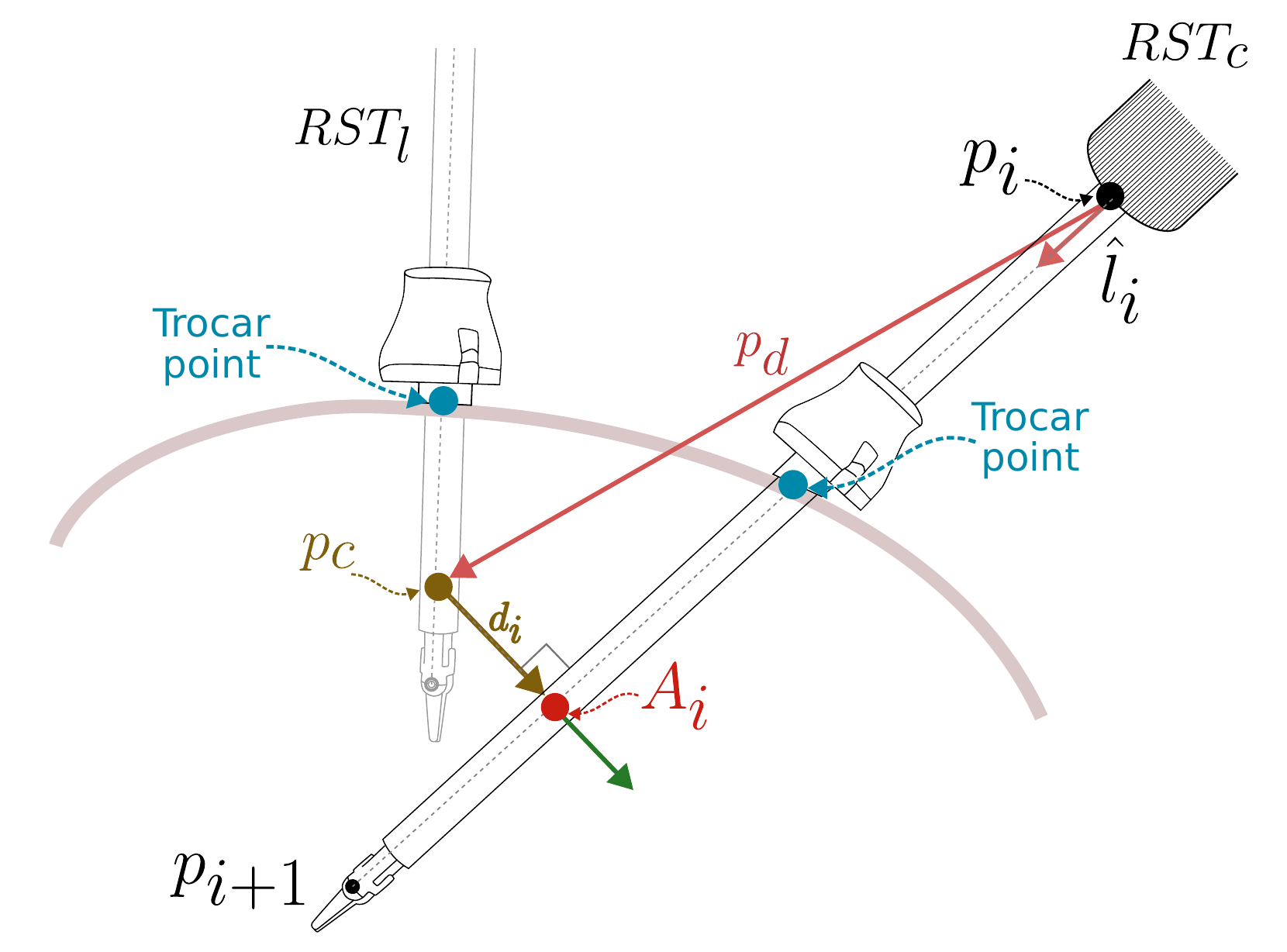}
  \caption{Characterization of the collision avoidance task.}
  \label{figurelabel}
\end{figure}

Given the presence of a static or dynamic obstacle located at $p_c \in \mathbb{R}^3$, we calculate the nearest point $p_{A_i} \in \mathbb{R}^3$ along the manipulator link $i$. This is done using the following.

\begin{equation}
    p_{A_i} = p_{i} + p_d^T \hat{l_i}\hat{l_i},
\end{equation}

where $\hat{l_i}=\frac{p_{i+1} - p_{i}}{||p_{i+1} - p_{i}||}$ indicates the direction of the surgical manipulator link axis and $p_d=p_{c}-p_{i}$ represents the difference between the position of the $i$th joint and the next joint of the manipulator. The vector $d_i = p_{A_i} - p_{c}$ denotes the vector from obstacle $p_c$ to the nearest point on the manipulator $i$th link $p_{A_i}$. 

By differentiating $||d_{i}||$ with respect to the manipulator joints, the Jacobian matrix of the collision task $ J_{coll}(q) \in \mathbb{R}^{1\times n}$ can be calculated as

\begin{equation}
\begin{aligned}
    J_{coll}(q) &= \frac{1}{2||d_i||} \frac{\delta d_i^T d_i}{\delta q} \\
    &=\hat{d_i}^T \left[ \left( I_{3} - \hat{l_i}\hat{l_i}^T \right)J_{i} + \left( \hat{l_i}p_d^T + p_d^T \hat{l_i}I_3 \right) \frac{\delta \hat{l_i}}{\delta q} \right]
\end{aligned}
\end{equation}
with
\begin{equation}
    \frac{\delta \hat{l_i}}{\delta q} = \frac{1}{||l_i||} \left( I_3 - \hat{l_i} \hat{l_i}^T \right) \left( J_{i+1} - J_i\  \right)
\end{equation}

where $J_i  \in \mathbb{R}^{3\times n}$ and $J_{i+1}  \in \mathbb{R}^{3\times n}$ are the Jacobian matrices of the $i$ link joint and the subsequent manipulator joint, respectively. The residual for the collision task $r_{coll}$ is defined as the minimum distance between the obstacle and the manipulator, and is given as 

\begin{equation}
    r_{coll} = ||d_i|| 
\end{equation}

\subsubsection{Manipulability maximization}
Maximizing the manipulability index $\mu$ \cite{yoshikawa85manipulability} can improve the performance of collision avoidance, providing larger ranges of motion and forces, and has previously been proposed for an HQP controller \cite{gholami21reconfigurable}. The manipulability maximization problem can be formulated as follows. 

\begin{equation}
    \min_{\dot{q}} \quad -m(q),
\end{equation}

where $m(q)=\sqrt{\det{JJ^T}}$. The equivalent optimization problem is presented in Eq. 14, and was proposed in \cite{dufour20maximizing}, where $\nabla m$ can be estimated numerically, resulting in faster computations suitable for real-time applications. 

\begin{equation}
    \min_{\dot{q}} \quad \frac{1}{2} \Delta t^2 \dot{q}^T \nabla m_k \nabla m_k^T \dot{q} + m \Delta t \nabla m^T \dot{q}
\end{equation}

\begin{figure}[t]
  \centering
  \includegraphics[width=0.9\linewidth]{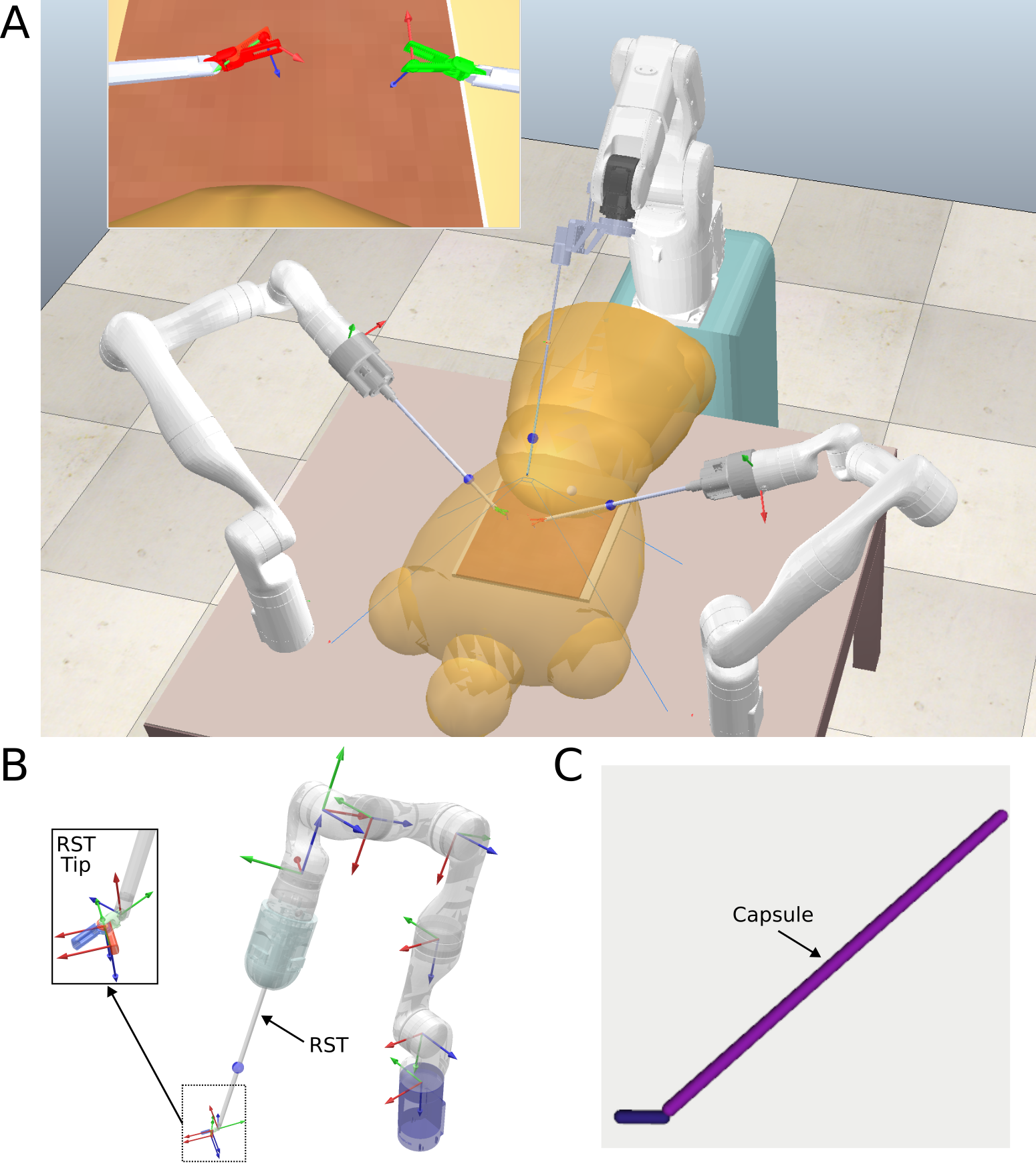}
  \caption{\textbf{A.} Simulation environment. Two surgical manipulators, introduce surgical robotic tools through trocar points. \textbf{B.} The kinematic chain of each surgical manipulator, comprising a 7-DOF robtic arm and  a  3-DOF robotic surgical tool (RST) \cite{colan23openrst}. \textbf{C.} Simplified collision model for each RST. The shaft and jaw links are represented as capsules.}
  \label{fig:4}
\end{figure}

\begin{figure}[t]
  \centering
  \includegraphics[width=\linewidth]{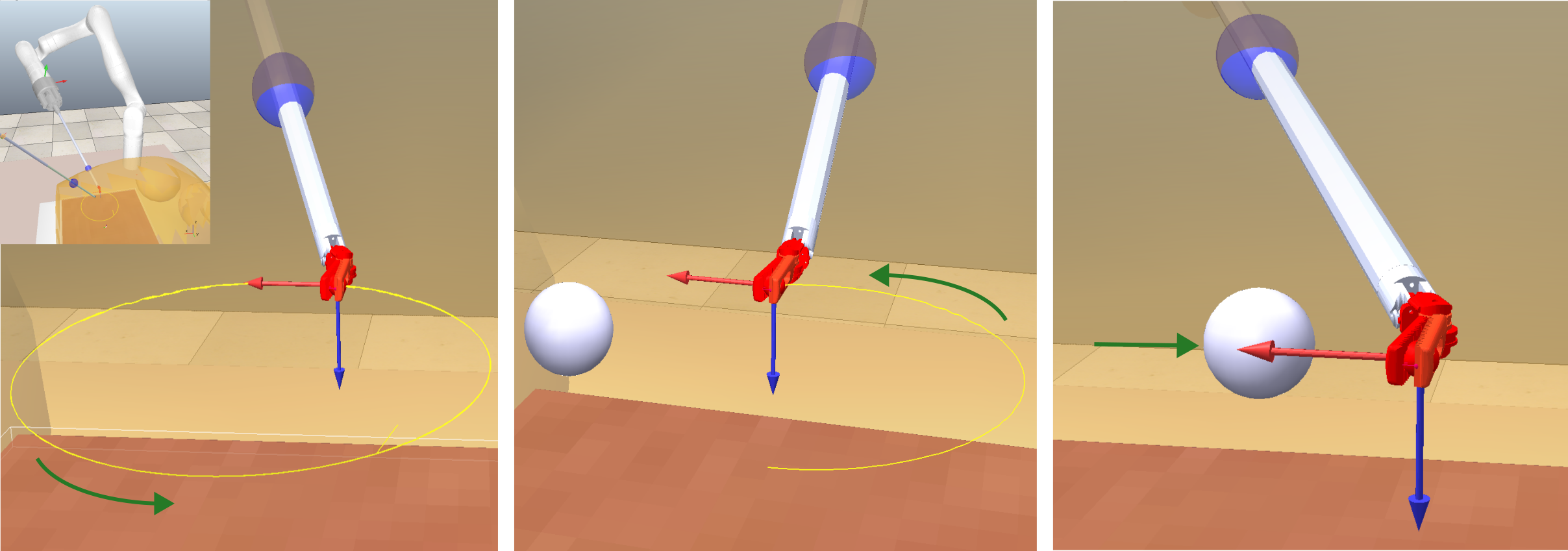}
  \caption{Visualization of the experimental cases. \textbf{A.} A 6D circular path tracking task without obstacles. \textbf{B.} Tracking task with a static object. \textbf{C.} Fixed position with a dynamic object.}
  \label{fig:5}
\end{figure}

The QP formulation for the manipulability maximization problem is then given by 

\begin{equation}
       \min_{\dot{q}} \quad  ||\Delta t \nabla m_k^T \dot{q} - m_k||_2^2
\end{equation}

\subsubsection{Joint limits}
The joint and velocity limits are defined as: 

\begin{equation}
\begin{aligned}
    q^- \le q \le q^+ \\
    -\dot{q}_{max} \le \dot{q} \le \dot{q}_{max}
\end{aligned}
\end{equation}

By adding convenient slack variables $w = [w^+ \; w^-]$, both constraints can be integrated into an optimization problem with inequality constraints given by

\begin{equation}
    \begin{aligned}
        \min_{\dot{q},w} \quad & \frac{1}{2} ||w||^2 \\
        \textrm{s.t.}  \quad & \quad  \dot{q}-\overline{q} \le w^+ \\
         & -\dot{q}+\underline{q} \le w^- 
    \end{aligned}
\end{equation}

where $\underline{q} = \max \left( \delta t (q^- -q_{act}), -\dot{q}_{max} \right)$ and $\overline{q} = \min \left( \delta t (q^+ -q_{act}), \dot{q}_{max} \right)$. The integration into the HQP controller is given by defining the variables $C$ and $d$ as
\begin{equation}
    C = \begin{bmatrix} I_n \\ -I_n \end{bmatrix}
    \quad\ \quad\ \quad\
    d = \begin{bmatrix} \overline{q} \\ -\underline{q} \end{bmatrix}
\end{equation}

\subsection{Smooth task priority transition}
The collision avoidance strategy should not be activated until an obstacle is recognized within a certain threshold distance $d_{\epsilon} = \epsilon_c + \alpha_c$. Once the criteria are met, the collision avoidance task takes a higher priority than the tracking task. To avoid instabilities from the transition between different task hierarchies, a transition gain $\beta_a$ is used, defined as 

\begin{equation}
    \beta_a = 1-\frac{||d_i||}{\epsilon_c}
\end{equation}

where $\epsilon_c$ is the maximum distance allowed from the manipulator and $\alpha_c$ is a tolerance buffer. The task weights for the Tracking task ($K_{t_{ee}} = 1-\beta_a$) and Collision avoidance task ($K_{t_{coll}} = \beta_a$) are updated to be a function of $\beta_a$, so the transition between both tasks is continuous.

\begin{figure}[t]
  \centering
  \includegraphics[width=0.7\linewidth]{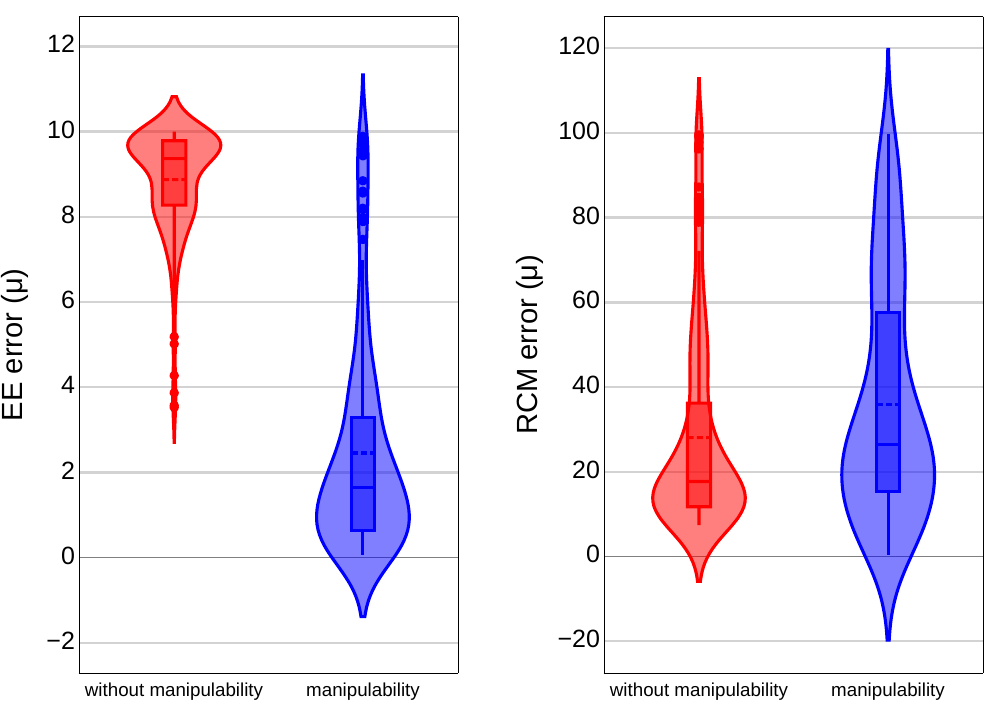}
  \caption{RCM and EE errors with and without considering the manipulability maximization task.}
  \label{fig:6}
\end{figure}

\begin{figure}[t]
  \centering
  \includegraphics[width=0.7\linewidth]{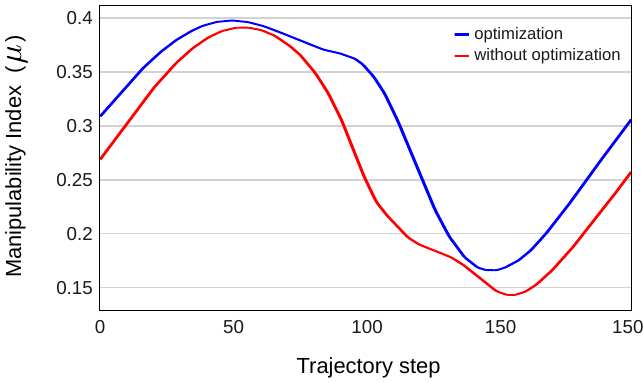}
  \caption{Manipulability index for a circular path tracking task. A comparison is presented including the manipulability maximization optimization as part of the stack of tasks. }
  \label{fig:7}
\end{figure}

\section{Experimental validation}

In our experimental validation, we implemented the HQP controller on a Linux Ubuntu 20.04 workstation equipped with an Intel Core i9-11900 processor and 64 GB of RAM. For kinematic computations, transformations, and kinematic chain parsing, we utilized the Pinocchio library (v. 2.6.10) \cite{carpentier19pinocchio}. The CASadi library (v. 3.5.5) \cite{andersson19casadi} served as a back-end for nonlinear and HQP solvers, while OSQP (v. 0.5.0) \cite{stellato20osqp} handled quadratic programming within the HQP solver, with warm start enabled. We conducted our experiments in the CoppeliaSim simulation environment \cite{rohmer13vrep}, as depicted in Fig.~\ref{fig:4}. We examined four distinct case scenarios, as illustrated in Fig.~\ref{fig:5}. Within the simulation environment, it is possible to obtain accurate positions of the forceps and collision objects. However, in real-world applications, this information can be estimated using real-time recognition and segmentation models \cite{fozilov23endoscope, davila24comparison}.

\begin{figure}[t]
  \centering
  \includegraphics[width=\linewidth]{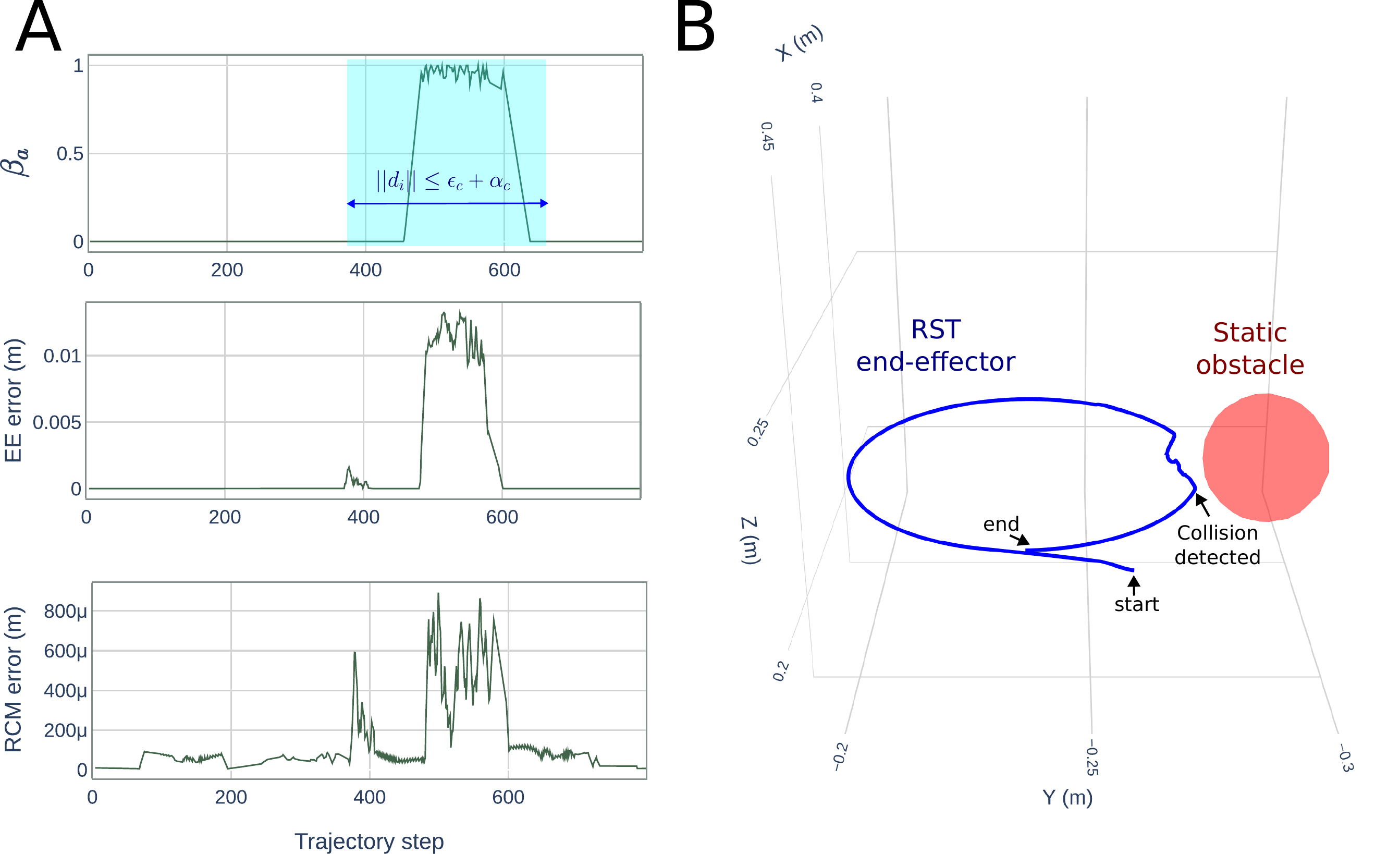}
  \caption{\textbf{A.} The transition gain, EE error and RCM errors. \textbf{B.} Trajectories followed by the end-effector with a static object.}
  \label{fig:8}
\end{figure}

\begin{figure}[t]
  \centering
  \includegraphics[width=\linewidth]{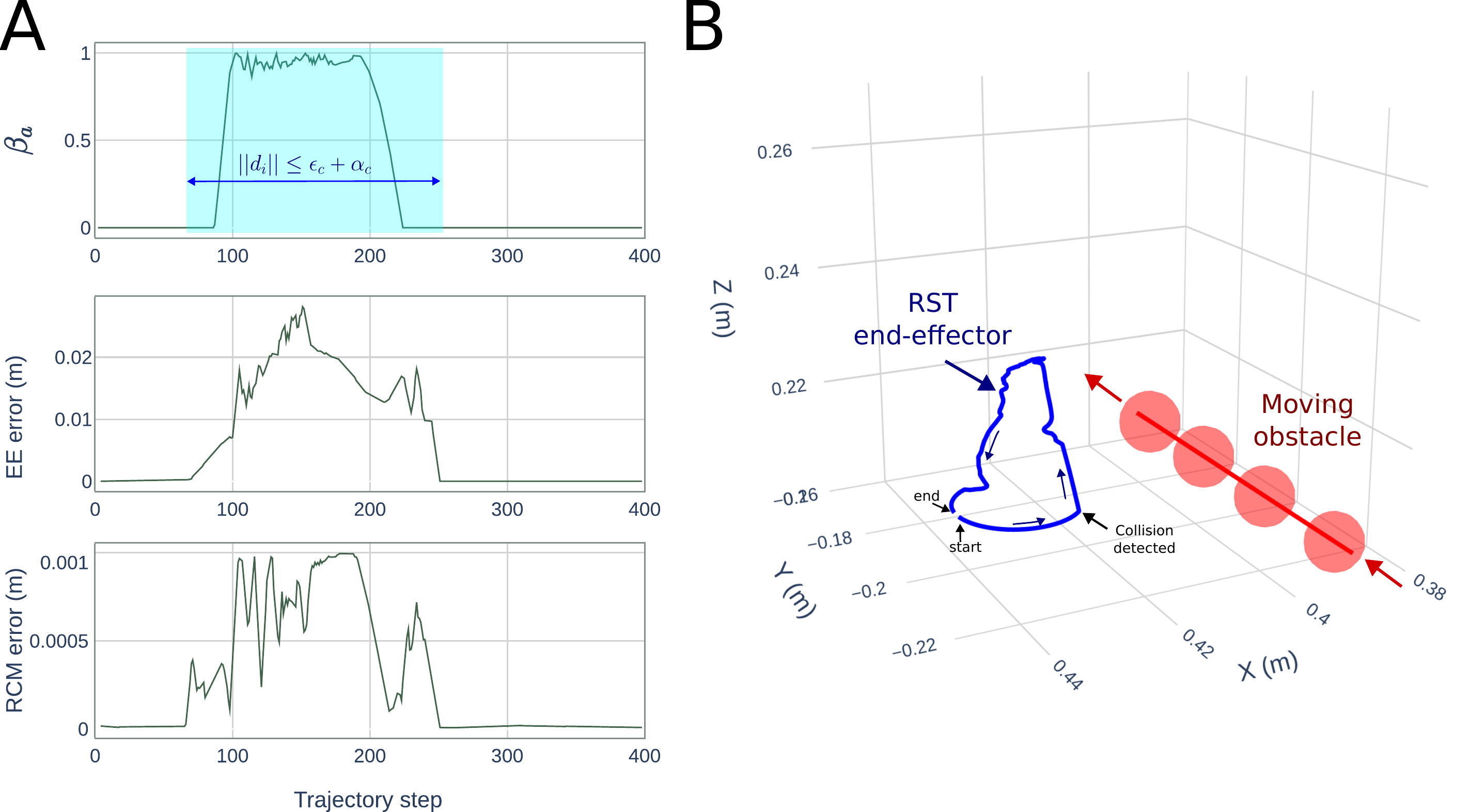}
  \caption{\textbf{A.} The transition gain, EE error and RCM errors. \textbf{B.} Trajectories followed by the end-effector with a dynamic object.}
  \label{fig:9}
\end{figure}

\subsection{Case I: No collision avoidance}

\begin{figure*}[h]
	\centering
	\includegraphics[width=7in]{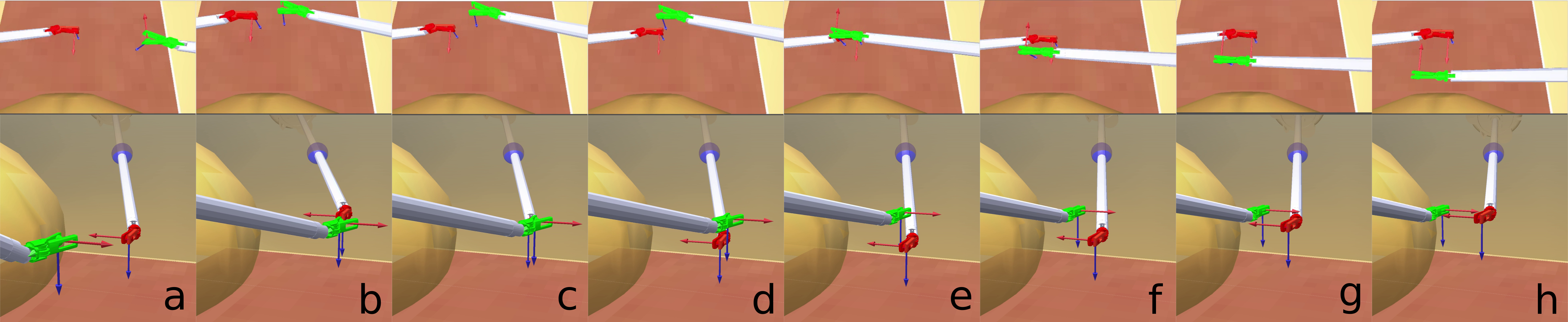}
	\caption{Snapshots of a collision avoidance performed when multiple surgical tools are involved.}
	\label{fig:10}
\end{figure*}

 In this scenario, we focus on the tracking and RCM tasks, with the RCM task given higher priority. We also introduced a Manipulability Maximization task and evaluated its performance. The results are summarized in Table~\ref{tab:1}, and Fig.~\ref{fig:6} shows the RCM and EE errors with and without manipulability maximization.

\begin{table}[h]
\caption{Performance metrics}
\label{tab:1}
\begin{center}
\vspace*{-5mm}
\begin{tabular}{|c||c|}
\hline
Performance metric & Value\\
\hline
Avg. EE pos. error ($10^{-6}$ m)& 2.45 $\pm$ 2.47\\
Max. EE pos. error ($10^{-6}$ m)& 9.88 \\
Avg. RCM error ($10^{-6}$ m) & 35.96 $\pm$ 27.29\\
Max. RCM error ($10^{-6}$ m) & 99.64\\
Avg. Manipulability index ($\mu$) (No max. opt.) & 0.266 $\pm$ 0.08\\
Avg. Manipulability index ($\mu$) (With max. opt.)& 0.303 $\pm$ 0.08\\
\hline
\end{tabular}
\end{center}
\end{table}

Figure~\ref{fig:6} shows the RCM and EE errors without and with maximization of manipulability. A reduction in the EE error is visible while the RCM error remains between the desired range ($e_{rcm}<10^{-4}$m). The evolution of the manipulability index is shown in Fig.~\ref{fig:7}, with the effect of the manipulability optimization visible, with a higher index in all trajectory steps.

\subsection{Case II: Collision avoidance with static object} 

The performance of the proposed framework in a collision avoidance task is evaluated when a static obstacle is involved. The surgical tool follows a circular path and a spherical obstacle is placed close to the given trajectory, as shown in Fig.~\ref{fig:5}B.  When the distance between the object and the surgical tool is within the given threshold ($d_\epsilon = 3 $ cm), the transition to a collision avoidance hierarchy starts, and the transition gain $\beta_a$ increases from 0 to 1. The trajectory followed by the tool is shown in Fig.~\ref{fig:8}B.

\vspace*{-1mm}
\begin{table}[h]
\caption{Performance metrics}
\label{table_example}
\begin{center}
\vspace*{-5mm}
\begin{tabular}{|c||c|}
\hline
Metric & Value\\
\hline
Max. EE pos. error (mm)& 0.33\\
Max. RCM error (mm) & 0.99\\
Avg. RCM error (mm) & 0.27 $\pm$ 0.33\\
Avg. Manipulability index ($\mu$) & 0.347\\
\hline
\end{tabular}
\vspace*{-5mm}
\end{center}
\end{table}

\subsection{Case III: Collision avoidance with dynamic object}
We performed a similar evaluation with a dynamic object moving toward the surgical tool, which maintains a fixed pose, as shown in Fig.~\ref{fig:5}C.  When the distance between the object and the surgical tool is less than $d_\epsilon$, the transition to a collision avoidance hierarchy begins, and the transition gain $\beta_a$ increases from 0 to 1 as shown in Fig.~\ref{fig:9}A. The trajectory followed by the tool is shown in Fig.~\ref{fig:9}B.

\begin{table}[h]
\caption{Performance metrics}
\label{table_example}
\begin{center}
\vspace*{-5mm}
\begin{tabular}{|c||c|}
\hline
Metric & Value\\
\hline
Max. EE pos. error (mm)& 0.49\\
Max. RCM error (mm) & 0.99\\
Avg. RCM error (mm) & 0.11 $\pm$ 0.20\\
Avg. Manipulability index ($\mu$) & 0.261\\
\hline
\end{tabular}
\vspace*{-5mm}
\end{center}
\end{table}

\subsection{Case IV: Collision avoidance between surgical tools}

Collision between two manipulators has been evaluated by giving each manipulator a circular path tracking task. The trajectory for both circular paths intersect, and at a given point, collision avoidance is activated. Snapshots of the motion of the tools are shown in Fig.~\ref{fig:10}.

\section{CONCLUSIONS}

This paper presents a hierarchical framework for collision avoidance in robot-assisted minimally invasive surgery. The proposed framework integrates multiple objectives, including maintaining the Remote Center of Motion (RCM) constraint, tracking tool poses, avoiding collisions, maximizing manipulability, and respecting kinematic limits. Experimental validation in simulated scenarios demonstrates the robustness and effectiveness of the proposed framework. It has been successfully tested in situations involving static and dynamic obstacles and collisions between surgical tools. Future work may involve further refinements and real-world testing of the framework to validate its performance. In addition, integration of sensory feedback and adaptive control strategies could contribute to even more sophisticated collision avoidance capabilities.

\addtolength{\textheight}{-11cm}   







\bibliographystyle{IEEEtran}
\bibliography{biblio}

\end{document}